\documentclass{article}

\usepackage[preprint,nonatbib]{nips_2018}

\usepackage[labelfont=bf]{caption}
\usepackage{textcomp}
\usepackage{graphicx}
\usepackage[utf8]{inputenc} % allow utf-8 input
\usepackage[T1]{fontenc}    % use 8-bit T1 fonts
\usepackage{url}            % simple URL typesetting
\usepackage{booktabs}       % professional-quality tables
\usepackage{amsfonts}       % blackboard math symbols
\usepackage{amsmath}
\usepackage{nicefrac}       % compact symbols for 1/2, etc.
\usepackage{microtype}      % microtypography
\usepackage{hyperref}       % hyperlinks

\title{Deep learning long-range information in undirected graphs with wave networks}

\author{
  Matthew K. Matlock\\
  \texttt{mmatlock@wustl.edu}\\
  \And
  Arghya Datta\\
  \texttt{arghya@wustl.edu}\\
  \And
  Na Le Dang\\
  \texttt{dangnl@wustl.edu}\\
  \And
  Kevin Jiang\\
  \texttt{jiangckevin@gmail.com}\\
  \And
  S. Joshua Swamidass\\
  \texttt{swamidass@wustl.edu}\\
  \AND
  \\
  Department of Pathology and Immunology\\
  Washington University in Saint Louis\\
  Saint Louis, MO 63139\\
}

\begin{document}

\maketitle

\begin{abstract}

    Graph algorithms are key tools in many fields of science and technology. Some of these algorithms depend on propagating information between distant nodes in a graph. Recently, there have been a number of deep learning architectures proposed to learn on undirected graphs. However, most of these architectures aggregate information in the local neighborhood of a node, and therefore they may not be capable of efficiently propagating long-range information. To solve this problem we examine a recently proposed architecture, wave, which propagates information back and forth across an undirected graph in waves of nonlinear computation. We compare wave to graph convolution, an architecture based on local aggregation, and find that wave learns three different graph-based tasks with greater efficiency and accuracy. These three tasks include (1) labeling a path connecting two nodes in a graph, (2) solving a maze presented as an image, and (3) computing voltages in a circuit. These tasks range from trivial to very difficult, but wave can extrapolate from small training examples to much larger testing examples. These results show that wave may be able to efficiently solve a wide range of problems that require long-range information propagation across undirected graphs. An implementation of the wave network, and example code for the maze problem are included in the tflon deep learning toolkit (\url{https://bitbucket.org/mkmatlock/tflon}).

\end{abstract}

\section{Introduction}

Deep learning is a powerful approach to machine learning in domains with complex data types~\cite{lecun2015deep, Schmidhuber2015, Baldi2003, Gers2000}. Modeling relationships among objects is a key analytic task in many fields. These relationships give rise to a natural graph data structure, where objects are nodes and relationships are edges, each of which can have properties. Recently, deep learning architectures, usually variants of convolutional neural networks, have been adapted for graph data~\cite{duvenaud2015convolutional, Kearnes2016}. Convolutional neural networks have become a standard tool for machine learning in euclidean domains, such as images~\cite{krizhevsky2012imagenet, ronneberger2015u}, videos~\cite{karpathy2014large}, and 3-d volumetric data~\cite{milletari2016v, pereira2016brain}. These approaches rely upon the key assumption that nodes can be efficiently described by local aggregation, either by summarizing the local graph topology, or by message passing between adjacent nodes. A recent review of these approaches acknowledges that long-range information is tacitly ignored~\cite{Bronstein2017}. Graph convolution only aggregates information locally; it does not efficiently propagate long-range information across a graph.

Algorithms operating on undirected graphs are key tools in nearly every field of science and technology. Notable examples include internet search~\cite{page1998pagerank}, bibliometrics~\cite{shibata2008detecting, ding2009pagerank}, social sciences~\cite{scott2017social}, cell biology~\cite{han2004evidence, Szklarczyk2014}, genomics~\cite{zerbino2008velvet}, chemistry~\cite{o2011open, raymond2002maximum}, circuit design~\cite{lengauer2012combinatorial}, and transportation engineering~\cite{magnanti1984network}. Some of these algorithms produce output for every edge or node, and small changes to the graph can have far-reaching impact on the output. For example, long-range information is required to determine whether nodes are in a cycle. Efficient algorithms for ring detection are well known~\cite{o2011open}; local information alone, however, is not sufficient to solve this problem. For many algorithms commonly used on graph data, it is difficult to intuit a solution based only on local aggregation, because information must be propagated across the whole graph.

In this study, we compare wave, an architecture designed to efficiently propagate information across a graph, with graph convolution. Wave has recently been shown to outperform graph convolution in modeling aromatic and conjugate system size, two fundamentally important graph algorithms in chemical informatics~\cite{Matlock2018}. Expanding upon that work, we show that wave outperforms graph convolution on three tasks: labeling a path connecting two nodes in a graph, (2) solving a maze presented as an image, and (3) computing voltages in a circuit. These three tasks range in difficulty, from trivial to very difficult. Wave achieves better accuracy and efficiency than graph convolution on all of these tasks, suggesting that wave might have value in a large range of problems where undirected graphs are used to represent data.

\section{Related work}

Techniques for representing graph structured data for input to deep learning architectures can be broadly separated into three classes: hand engineered features, unsupervised learning, and supervised learning. These approaches almost exclusively focus on finding node features that characterize local topology. 

Historically, hand engineered features were commonly used because techniques for automated representation learning did not exist. Hand engineered features describe the local topology near a node in terms of its degree, and the connectivity properties of subgraphs formed by extracting a node, its neighbors, and their collected edges \cite{gallagher2010leveraging, henderson2011s, zaretzki2013xenosite, Matlock2015}.

Several unsupervised techniques for graphs utilize random walks on local neighborhoods of a node to generate data for learning fixed-length vector embeddings. DeepWalk~\cite{Perozzi2014} adapts the word2vec~\cite{mikolov2013efficient} approach for word embeddings by modeling nodes as words and random walks as sentences. LINE~\cite{tang2015line}, extends the random walk approach by introducing an objective that preserves properties of the local topology. Node2vec~\cite{Grover2016} adds a more flexible definition of node neighborhoods realized by random walks. These unsupervised embedding approaches are closely related to graph kernels, which define a similarity metric between nodes that can be used to impute labels on similar nodes~\cite{hofmann2008kernel, vishwanathan2010graph}.

Supervised approaches utilize custom neural network components that can operate directly on graphs. Traditional convolutional neural networks require uniform spatial or temporal sampling of data, while graphs have flexible spatial structure. Several works generate receptive fields for traditional neurons by aggregating node and edge features over local neighborhoods~\cite{niepert2016learning, duvenaud2015convolutional}. Extending upon these approaches, specialized graph convolution architectures have also been proposed~\cite{li2015gated, kipf2016semi, Kearnes2016}. These architectures use learnable message functions, which allow nodes to communicate with one another along their shared edges~\cite{gilmer2017neural}.

Two key works have introduced recurrent neural networks operating on graphs. Tree LSTMs operate on trees (directed graphs without cycles), aggregating data from multiple child nodes to a parent node. Tree LSTMs achieve state of the art performance on language modeling tasks~\cite{tai2015improved}. Inner recursive neural networks operate on undirected graphs by enumerating a unique spanning tree for each node and applying an architecture similar to Tree LSTMs. These networks have shown promise in several chemical informatics tasks~\cite{baldi2018inner}. Unfortunately, neither of these techniques can operate on undirected graphs with cycles without loss of information.

In addition to graph specific architectures a general purpose architecture, the differentiable neural computer~\cite{graves2014neural, graves2016hybrid}, can enumerate paths connecting nodes in a graph. Unfortunately, this architecture must be trained by reinforcement learning, a huge computational cost, and it does not scale efficiently to very large graphs.

\section{Graph convolution networks}

\begin{figure}
  \centering
  \includegraphics[width=5.5in]{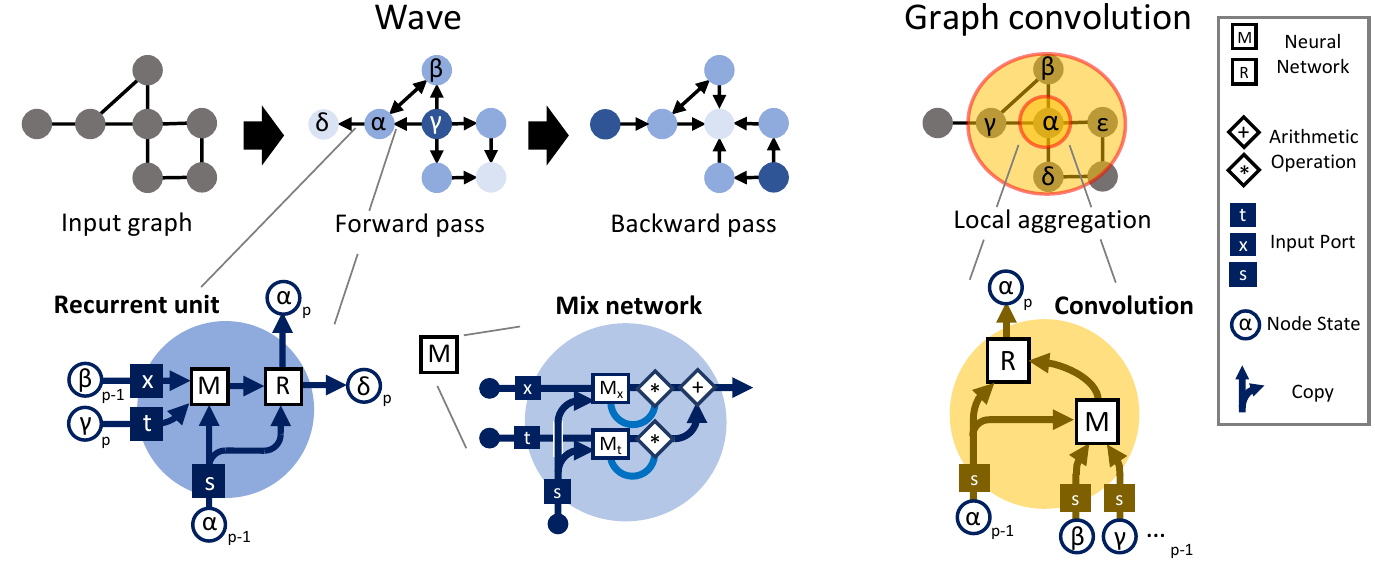}
    \caption{\textbf{Wave enables long-range information propagation in graph structured data.} Two alternative paradigms for computing node representations in graphs: wave (left, blue) and graph convolution (right, orange). Graph convolution is a member of a general class of methods based on local aggregation of information. For each convolutional pass, updating a node depends on the node itself and its neighbors. Sufficient passes must be computed to ensure all nodes share information. In contrast, wave passes information forward and backward across a graph in a single pass. To achieve this, nodes and edges are ordered by breadth first search, starting from a central node. Each node update depends upon its immediate ancestors, and is propagated to descendant nodes.}
  \label{fig:methods}
\end{figure}

Graph convolution has shown promising results in subgraph isomorphism~\cite{scarselli2009graph}, program verification~\cite{li2015gated}, chemical informatics~\cite{Kearnes2016, duvenaud2015convolutional}, and quantum chemistry~\cite{gilmer2017neural}. Graph convolution learns node representations by repeated aggregation over local neighborhoods (Figure~\ref{fig:methods}). The state update on pass $p$ for a node $u$ with neighbors $v \in N$ is defined as:
\begin{equation}
    S_u^p  = R_1^p(S_u^{p-1}, M_u^p),
    \label{eq:state_update}
\end{equation}
where $R_1^p$ is a recurrent neural network, which is parameterized differently for each pass, $S_u^{p-1}$ is the node state from the prior pass, and $M_u^p$ is a message function, defined by,
\begin{equation}
    M_u^p = \sum_{v \in N} c_1^p(E_{uv}^{p-1}, S_v^{p-1}),
\end{equation}
where $c_1^p$ is a neural network, and $E_{uv}^{p-1}$ is the state of edge $u,v$ from the prior pass. The edge states are updated by the recurrence
\begin{equation}
    E_{uv}^p  = R_2^p(E_{uv}^{p-1}, M_{uv}^p).
\end{equation}
Once again, $M_{uv}^p$ is the output of a message function, defined by
\begin{equation}
    M_{uv}^p  = c_2^p(S_u^{p-1}, S_v^{p-1}) + c_2^p(S_v^{p-1}, S_u^{p-1}),
\end{equation}
where $c_2^p$ is another neural network.
In this work, the $R_*^p$, and $c_*^p$ are each dense layers with exponential linear (ELU) activation~\cite{clevert2015fast}. Additional details can be found in Kearnes \textit{et al}~\cite{Kearnes2016}.
Graph convolution requires as many passes as the radius of a graph to propagate information between all nodes~\cite{Matlock2018}.

\section{Wave networks}

In wave, similar to graph convolution, node states are updated based on the prior node states and a message constructed from neighbor states (Equation~\ref{eq:state_update}). However, updates are ordered by breadth first search, and information is propagated from a central node in a wave across the entire graph (Figure~\ref{fig:methods}). Updates are executed in forward and backward passes. During a forward pass, the node update depends only upon ancestor nodes, and the order is reversed in the backward pass. This node ordering enables information to traverse the entire graph in a single pass. 

The wave message function is computed by a special mix network, which only uses information from incoming nodes $I$. Formally, given a node $u$ with incoming nodes $v \in I$, the message function $M_u^p$ for pass $p$ is,
\begin{equation}
    M_{u}^{p} = b + \sum_{v \in I} w \odot S_{u}^{p-1} \odot \frac{n_{1,\pi(u,v)}(S_{u}^{p-1}, E_{uv}, S_{v}^{p})}{\sum_{v \in I} n_{1,\pi(u,v)}(S_{u}^{p-1}, E_{uv}, S_{v}^{p})} + S_{u}^{p-1} \odot n_{2,\pi(u,v)}(S_{u}^{p-1}, E_{uv}, S_{v}^{p}), \\
\end{equation}
where $\odot$ is element-wise vector multiplication, $\pi(u,v)$ is an index (0 or 1) chosen by the type of edge in the breadth first search (tree or cross), the $n_{1,\pi(u,v)}$ are neural networks with exponential activation, the $n_{2,\pi(u,v)}$ are neural networks with softsign activation~\cite{glorot2010understanding}, $w$ is a weight vector, and $b$ is a bias vector. For cross edges, the sibling node state from the previous pass is used, as the current pass state is not yet available. Informally, the mix network computes two weighted sums of the input states: one weighted by a neural network with softmax output and one weighted by a neural network with softsign output. The mix network can range between computing a weighted sum, a maximum, or a minimum of the input states, depending upon its parameters. Additional details of this architecture can be found in Matlock \textit{et al}~\cite{Matlock2018}. 

In Matlock \textit{et al}~\cite{Matlock2018}, the default recurrent unit ($R^p$) for wave was a variant of the gated recurrent unit (GRU)~\cite{jozefowicz2015empirical, cho2014properties}, which we refer to as miniGRU. Briefly, a GRU is a neural memory cell consisting of three gates: read, update, and output. The miniGRU differs by omitting the read gate, which is redundant with the mix network. A formal description is available in the supplementary methods.

\section{Finding paths in trees and graphs}

\begin{figure}
  \centering
  \includegraphics[width=5.5in]{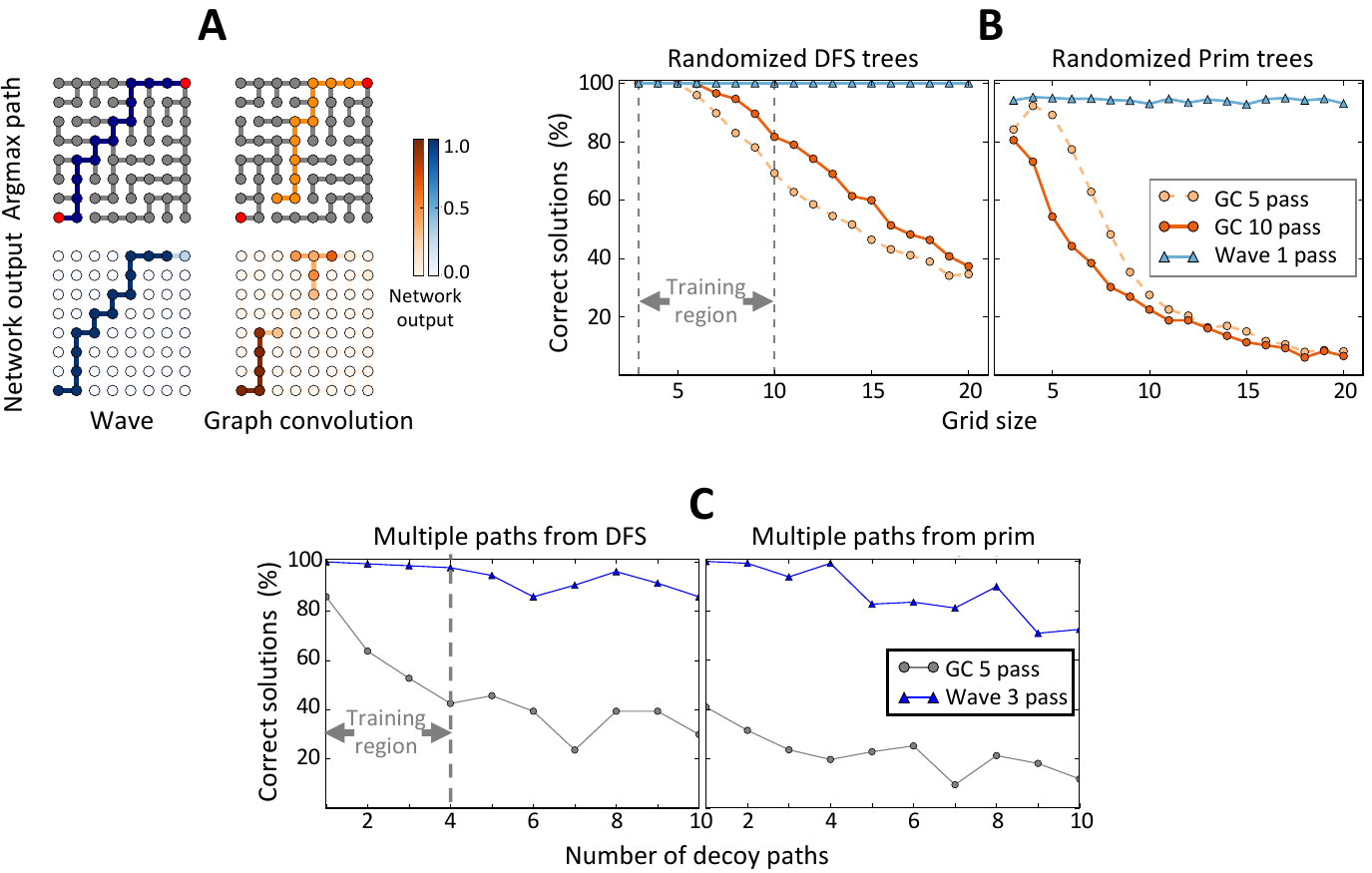}
    \caption{\textbf{Wave networks label paths between distant nodes in trees and graphs.} (A) Graph convolution assigns low-confidence scores to nodes distant from the goals resulting in errors, while wave assigns high confidence to all path nodes. (B) A single wave forward-backward pass is sufficient to perfectly solve path-finding in trees constructed by randomized depth-first search, and generalizes to larger trees and to trees generated with a randomized Prim algorithm. In contrast, graph convolutional methods are unable to learn even the training data accurately. (C) In addition, multiple passes of wave can determine the shortest path in graphs with varying numbers of paths between the start and goal. Critically, wave models trained only on examples with fewer potential paths can extrapolate to mazes with far more potential paths.}
  \label{fig:maze_graphs}
\end{figure}

Labeling nodes in a path between two other nodes requires propagating long-range information. This problem is easily solved with a depth-first search; however, wave propagates information in a breath-first search, which does not start at either of the goal nodes. Nevertheless, we expected wave to outperform graph convolution in this task. To test this hypothesis, we constructed random path finding examples. For each example, a spanning tree was constructed from a grid graph via either a randomized depth first search (DFS) or a randomized Prim's algorithm~\cite{kleinberg2006algorithm}; edges in the spanning tree were retained as an undirected graph. Each network was provided the same features (the goal nodes), and the same output layer (dense layer with sigmoid activation). For this experiment, wave used a dense layer with $\tanh$ activation for its recurrent network. Wave used a node state size of $10$. Graph convolution used a node state size of $5$ and edge state size of $5$. We trained both networks to label nodes traversed by the correct path using the cross entropy loss~\cite{shore1981properties} with the Adam optimizer~\cite{kingma2014adam}. Optimizer parameters were: learning rate $1 \times 10^{-3}$, mini-batch size $50$, and $30,000$ iterations. Models were implemented in the Tensorflow deep learning framework~\cite{abadi2016tensorflow}. Networks were trained with a curriculum learning scheme~\cite{bengio2009curriculum}, starting with small graphs of size three ($3 \times 3$ nodes) up to size $10$ (supplementary methods). Larger examples were introduced every $1500$ iterations. Solutions were marked correct if the labeled path could be enumerated by starting at a goal node and transitioning to the adjacent node with maximum probability without backtracking (argmax solution).

A single wave forward-backward pass perfectly solved the DFS path finding problem, both on trees of similar size as the training domain, and on trees much larger than those in the training data (Figure~\ref{fig:maze_graphs}B). Furthermore, the solution learned by the wave network was generalized enough to achieve high performance on the Prim path finding problem. In contrast, graph convolution with $5$ or $10$ passes was not able to solve the DFS problem. Graph convolution with $10$ passes performed worse on the Prim path finding problem, suggesting that the solutions found by graph convolution may overfit the data. Examining the network outputs for an example from the Prim test, we see that while the wave network clearly marks path nodes, the graph convolutional network exhibits uncertainty in marking nodes that are far from a goal node (Figure~\ref{fig:maze_graphs}A). In addition, the wave network had fewer parameters than the $10$ pass graph convolution network ($1641$ vs. $2576$). These findings support our hypothesis that wave propagates information more efficiently than graph convolution.

In addition to trees, a wave network with three forward-backward passes was able to find the shortest path in graphs with multiple possible paths between the start and end goal (Figure~\ref{fig:maze_graphs}C). For this problem, we generated mazes with multiple possible paths by starting with a DFS (or Prim) tree from the previous task, and connecting randomly chosen pairs of adjacent nodes in the grid. The resulting graphs were categorized by counting the number of resulting paths. Critically, wave was able to extrapolate from graphs with a few possible paths (one to four) to find shortest paths in graphs with many more possible paths (five to ten). Furthermore, wave was able generalize from mazes generated by the DFS algorithm to mazes generated with the Prim algorithm. In contrast, graph convolution with $5$ passes was not able to achieve the same performance as wave on the training set, did not extrapolate well to the test data, and did not generalize well to mazes generated with the Prim algorithm

\section{Solving a maze in an image}

\begin{figure}
  \centering
  \includegraphics[width=5.5in]{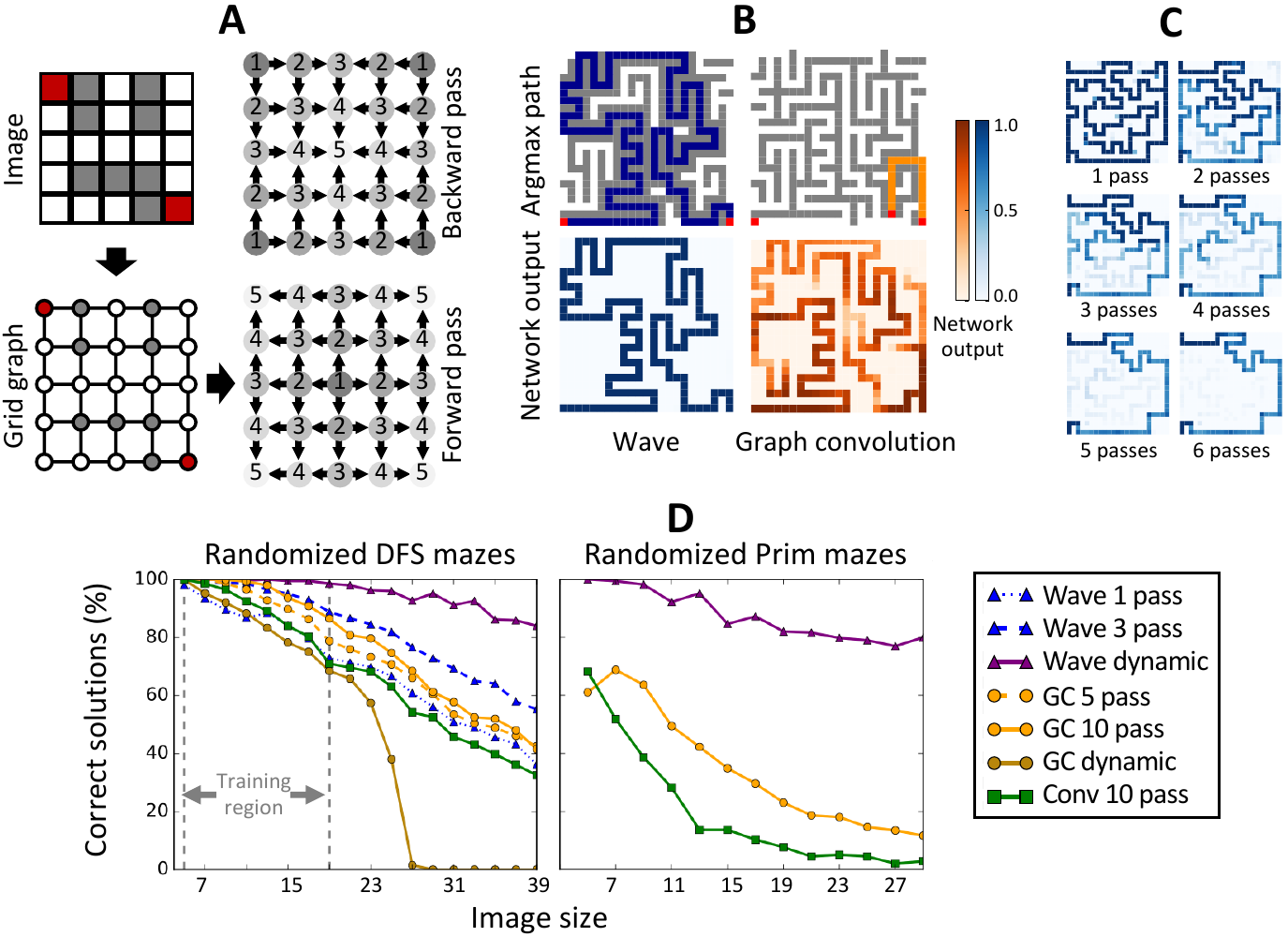}
    \caption{\textbf{Wave networks solve mazes presented as images.} (A) We construct images of randomly generated mazes with labeled walls, start and end points, and converted these to grid graphs where each node represents a pixel and adjacent pixels are connected by edges. Wave processes the image in expanding and contracting shells. (B) Wave can find the correct path connecting the start and endpoints. In contrast, convolutional approaches label many decoy paths. (C) Dynamic wave learns a progressive erasing algorithm, where each successive pass removes weight from decoy paths. (D) Wave outperforms both graph convolution and image convolution with max pooling. A dynamic version of the wave, which adapts the number of passes to the image size, substantially outperforms all other tested methods and generalizes to randomized Prim mazes.}
  \label{fig:maze_images}
\end{figure}

Solving mazes represented as images is much more difficult than path finding in a tree, because long-range information must be explicitly routed along valid paths, while avoiding walls. We translated the DFS and Prim graphs from the previous task into maze images. To create an undirected graph, the images were translated into pixel grid graphs, where adjacent pixels are connected by edges (Figure~\ref{fig:maze_images}A). In these graphs, nodes are labeled as passable, wall or goal. Wave network messages propagate radially out from the center of the image, and are not restricted to the passable areas of the maze. The network must label the pixels that connect the two goals, without crossing a wall. Such a task is arguable more difficult for humans than image recognition; while we can tell at a glance that an image is a maze, solving a large maze requires meticulously tracing paths through the maze.

In addition to testing graph convolution and wave with fixed numbers of passes, we also introduced dynamic versions of these architectures, which perform $(N+1)/2$ passes for images of size $N \times N$. In the dynamic versions of the architectures, weights are shared between each pass, which substantially reduces the overall number of parameters. Finally, we also tested image convolutional neural networks with $3 \times 3 \times 16$ kernels, ELU activations, and max pooling at each pass. The network specifications and training procedures used for this problem are identical to those from the previous section, except that models were trained for $60,000$ iterations. 

Wave with three or more passes substantially outperformed graph convolution and image convolution on both DFS and Prim mazes (Figure~\ref{fig:maze_images}D). Examining network outputs showed that, while the dynamic wave was able to accurately label path nodes with high probability, graph convolution labeled many nodes outside of the correct path (Figure~\ref{fig:maze_images}B). Using the dynamic wave, we generated outputs for varying numbers of passes. The algorithm learned by this network appears to progressively erase portions of the maze that are not part of the correct path, eventually converging when no non-path elements remain labeled (Figure~\ref{fig:maze_images}C). We also note that the dynamic wave has the fewest parameters ($1661$) of all tested networks, with the exception of the dynamic graph convolution. These findings provide evidence that wave can propagate long-range information in images.

\section{Computing voltages in a circuit}

\begin{figure}
  \centering
  \includegraphics[width=5.5in]{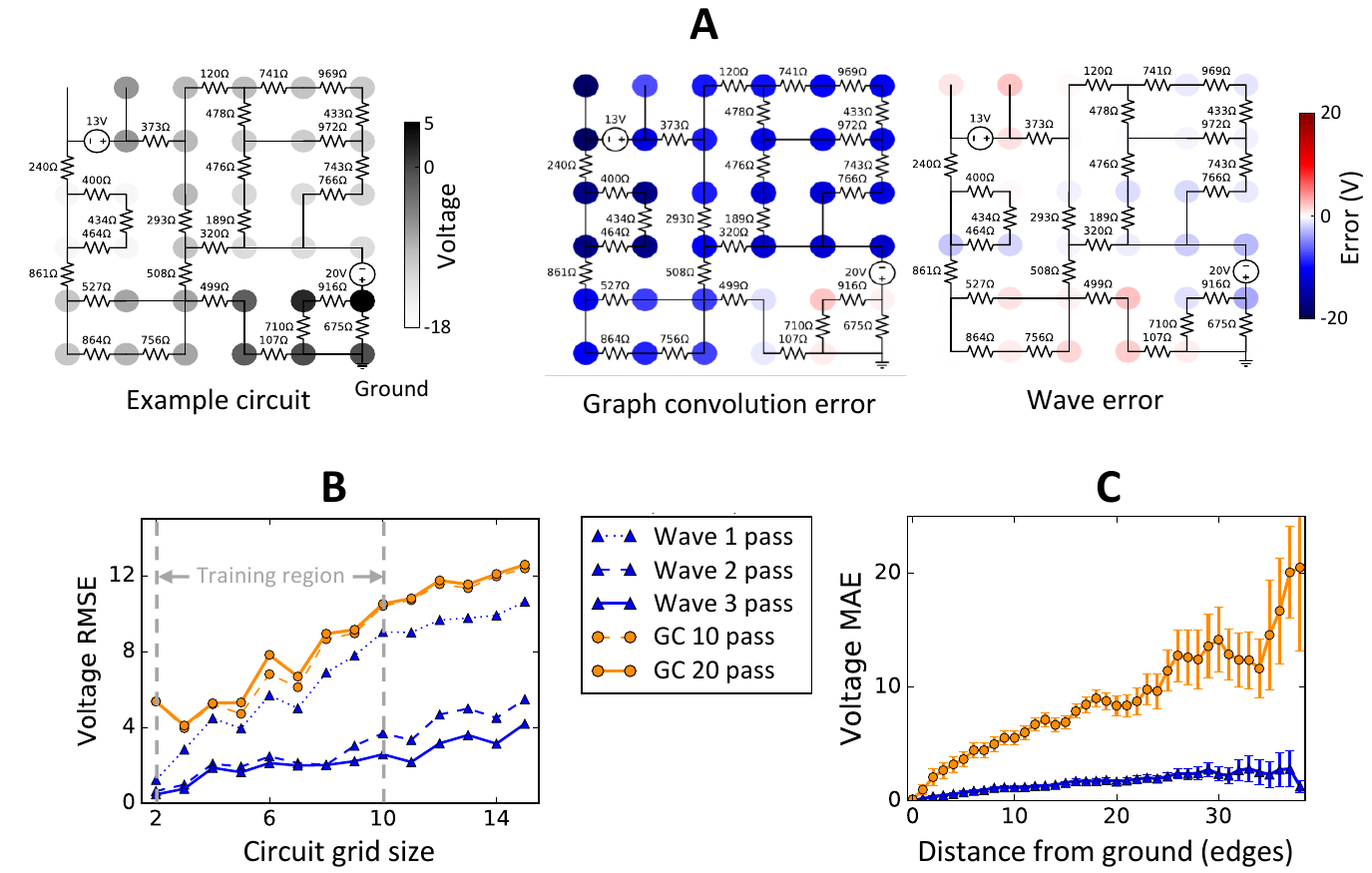}
    \caption{\textbf{Wave networks compute voltages in a circuit.} (A) We constructed randomized circuits consisting of a single ground node (lower right), and randomly selected components (batteries, resistors and wires). The task was to label each junction node with its voltage compared to the ground node. Wave exhibited low error rates compared to graph convolutional approaches. (B) Graph convolution root-mean-square error (RMSE) increased quickly with circuit size compared to wave with more than one pass. Additional convolutional passes did not improve performance on this task. (C) Graph convolution exhibits increasing mean-absolute error (MAE) with distance from the ground node.\label{fig:circuits}}
\end{figure}

Computing voltages in a circuit is a more complex graph problem than path finding in trees or images because it requires information propagation simultaneously between many interdependent nodes. The voltage at a node depends not only on its relationship with the ground node, but also on paths passing through batteries. We generated random circuits from grid graphs by randomly deleting edges, and replacing the remaining edges with one of three component types (wires, resistors or batteries) with randomly chosen properties (Figure~\ref{fig:circuits}A, supplementary methods). Nodes connected to a battery's positive terminal were labeled to indicate battery orientation. One node was labeled ground. Voltages were calculated using PySpice~\cite{PySpice}. Wave parameters were: state size $20$, miniGRU recurrent unit. Graph convolution parameters were: node state size $15$, edge state size $5$. Training parameters were the same as previous problems, except mini-batch size decreased with increasing example size to accommodate circuit simulator runtime (supplementary methods).

All tested wave networks achieved lower root-mean-square error compared to graph convolutional (Figure~\ref{fig:circuits}B). Wave with two or more passes exhibit a substantially slower growth in error rates with circuit size, even for circuits larger than those presented during training. Interestingly, increasing the number of convolutional passes did not improve the performance of graph convolution on this task. An individual circuit example reveals that errors made by graph convolution calculations are substantially biased for nodes far from the ground node, while wave errors are small, and unbiased (Figure~\ref{fig:circuits}A). In addition, when examining circuits of size $10$, graph convolution errors rose in proportion with target node distance from the ground node, while wave errors exhibited substantially lower correlation with distance (Figure~\ref{fig:circuits}C). The $1$ pass wave had substantially fewer parameters than the best graph convolution (12261 vs 20376), but still outperformed graph convolution on all tested examples sizes. These findings provide evidence that wave can propagate information among multiple, interdependent nodes over long ranges more efficiently than graph convolution.

\section{Conclusion}

Wave efficiently represents long-range dependencies in graphs for several different tasks when compared to graph convolution. Wave can (1) label paths connecting distant vertices in undirected graphs, (2) label paths in mazes represented as images, and (3) compute voltage potentials in circuits represented as graphs. Furthermore, wave can extrapolate from small training examples to much larger testing examples. In all of these experiments, the top wave networks also had the fewest parameters. While this study only examined relatively small graphs (up to $400$ nodes), the computational and memory complexity of wave enable it to scale to much larger graphs. We are optimistic that wave may efficiently solve many problems where representing long-range dependencies in graphs is critical.

\subsection*{Acknowledgments}

The authors declare that they have no competing financial conflicts of interests.
The authors are also grateful to the developers of Tensorflow and the open-source cheminformatics tools OpenBabel and RDKit.
Research reported in this publication was supported by the National Library Of Medicine of the National Institutes of Health under Award Numbers R01LM012222 and R01LM012482, and by the National Institutes of Health under Award Number GM07200.
The content is the sole responsibility of the authors and does not necessarily represent the official views of the National Institutes of Health.
Computations were performed using the facilities of the Washington University Center for High Performance Computing, which were partially funded by NIH grants nos. 1S10RR022984-01A1 and 1S10OD018091-01.
We also thank both the Department of Immunology and Pathology at the Washington University School of Medicine, the Washington University Center for Biological Systems Engineering and the Washington University Medical Scientist Training Program for their generous support of this work.

\bibliographystyle{acm}
\bibliography{graph_deeplearning}

\begin{thebibliography}{10}

\bibitem{abadi2016tensorflow}
{\sc Abadi, M., Barham, P., Chen, J., Chen, Z., Davis, A., Dean, J., Devin, M.,
  Ghemawat, S., Irving, G., and Isard, M.}
\newblock Tensorflow: A system for large-scale machine learning.
\newblock In {\em Proceedings of the 12th USENIX conference on operating
  systems design and implementation\/} (2016), vol.~16, pp.~265--283.

\bibitem{baldi2018inner}
{\sc Baldi, P.}
\newblock The inner and outer approaches to the design of recursive neural
  architectures.
\newblock {\em Data mining and knowledge discovery 32}, 1 (2018), 218--230.

\bibitem{Baldi2003}
{\sc Baldi, P., and Pollastri, G.}
\newblock {The Principled Design of Large-Scale Recursive Neural Network
  Architectures--DAG-RNNs and the Protein Structure Prediction Problem}.
\newblock {\em Journal of Machine Learning Research 4}, Sep (2003), 575--602.

\bibitem{bengio2009curriculum}
{\sc Bengio, Y., Louradour, J., Collobert, R., and Weston, J.}
\newblock Curriculum learning.
\newblock In {\em Proceedings of the 26th annual international conference on
  machine learning\/} (2009), ACM, pp.~41--48.

\bibitem{Bronstein2017}
{\sc Bronstein, M.~M., Bruna, J., LeCun, Y., Szlam, A., and Vandergheynst, P.}
\newblock Geometric deep learning: Going beyond euclidean data.
\newblock {\em {IEEE} Signal Processing Magazine 34}, 4 (jul 2017), 18--42.

\bibitem{cho2014properties}
{\sc Cho, K., van Merrienboer, B., Bahdanau, D., and Bengio, Y.}
\newblock On the properties of neural machine translation: Encoder--decoder
  approaches.
\newblock In {\em Proceedings of SSST-8, Eighth Workshop on Syntax, Semantics
  and Structure in Statistical Translation\/} (2014), Association for
  Computational Linguistics, pp.~103--111.

\bibitem{clevert2015fast}
{\sc Clevert, D.-A., Unterthiner, T., and Hochreiter, S.}
\newblock Fast and accurate deep network learning by exponential linear units
  (elus).
\newblock {\em arXiv preprint arXiv:1511.07289\/} (2015).

\bibitem{ding2009pagerank}
{\sc Ding, Y., Yan, E., Frazho, A., and Caverlee, J.}
\newblock Pagerank for ranking authors in co-citation networks.
\newblock {\em Journal of the Association for Information Science and
  Technology 60}, 11 (2009), 2229--2243.

\bibitem{duvenaud2015convolutional}
{\sc Duvenaud, D.~K., Maclaurin, D., Iparraguirre, J., Bombarell, R., Hirzel,
  T., Aspuru-Guzik, A., and Adams, R.~P.}
\newblock Convolutional networks on graphs for learning molecular fingerprints.
\newblock In {\em Advances in neural information processing systems\/} (2015),
  pp.~2224--2232.

\bibitem{gallagher2010leveraging}
{\sc Gallagher, B., and Eliassi-Rad, T.}
\newblock Leveraging label-independent features for classification in sparsely
  labeled networks: An empirical study.
\newblock In {\em Advances in Social Network Mining and Analysis}. Springer,
  2010, pp.~1--19.

\bibitem{Gers2000}
{\sc Gers, F.~A., Schmidhuber, J., and Cummins, F.}
\newblock {Learning to Forget: Continual Prediction with LSTM}.
\newblock {\em Neural Computation 12}, 10 (oct 2000), 2451--2471.

\bibitem{gilmer2017neural}
{\sc Gilmer, J., Schoenholz, S.~S., Riley, P.~F., Vinyals, O., and Dahl, G.~E.}
\newblock Neural message passing for quantum chemistry.
\newblock {\em arXiv preprint arXiv:1704.01212\/} (2017).

\bibitem{glorot2010understanding}
{\sc Glorot, X., and Bengio, Y.}
\newblock Understanding the difficulty of training deep feedforward neural
  networks.
\newblock In {\em Proceedings of the thirteenth international conference on
  artificial intelligence and statistics\/} (2010), pp.~249--256.

\bibitem{graves2014neural}
{\sc Graves, A., Wayne, G., and Danihelka, I.}
\newblock Neural turing machines.
\newblock {\em arXiv preprint arXiv:1410.5401\/} (2014).

\bibitem{graves2016hybrid}
{\sc Graves, A., Wayne, G., Reynolds, M., Harley, T., Danihelka, I.,
  Grabska-Barwi{\'n}ska, A., Colmenarejo, S.~G., Grefenstette, E., Ramalho, T.,
  Agapiou, J., et~al.}
\newblock Hybrid computing using a neural network with dynamic external memory.
\newblock {\em Nature 538}, 7626 (2016), 471.

\bibitem{Grover2016}
{\sc Grover, A., and Leskovec, J.}
\newblock node2vec: Scalable feature learning for networks.
\newblock In {\em Proceedings of the 22nd {ACM} {SIGKDD} International
  Conference on Knowledge Discovery and Data Mining - {KDD}
  {\textquotesingle}16\/} (2016), {ACM} Press.

\bibitem{han2004evidence}
{\sc Han, J.-D.~J., Bertin, N., Hao, T., Goldberg, D.~S., Berriz, G.~F., Zhang,
  L.~V., Dupuy, D., Walhout, A.~J., Cusick, M.~E., Roth, F.~P., et~al.}
\newblock Evidence for dynamically organized modularity in the yeast
  protein--protein interaction network.
\newblock {\em Nature 430}, 6995 (2004), 88.

\bibitem{henderson2011s}
{\sc Henderson, K., Gallagher, B., Li, L., Akoglu, L., Eliassi-Rad, T., Tong,
  H., and Faloutsos, C.}
\newblock It's who you know: graph mining using recursive structural features.
\newblock In {\em Proceedings of the 17th ACM SIGKDD international conference
  on knowledge discovery and data mining\/} (2011), ACM, pp.~663--671.

\bibitem{hofmann2008kernel}
{\sc Hofmann, T., Sch{\"o}lkopf, B., and Smola, A.~J.}
\newblock Kernel methods in machine learning.
\newblock {\em The annals of statistics\/} (2008), 1171--1220.

\bibitem{jozefowicz2015empirical}
{\sc Jozefowicz, R., Zaremba, W., and Sutskever, I.}
\newblock An empirical exploration of recurrent network architectures.
\newblock In {\em International Conference on Machine Learning\/} (2015),
  pp.~2342--2350.

\bibitem{karpathy2014large}
{\sc Karpathy, A., Toderici, G., Shetty, S., Leung, T., Sukthankar, R., and
  Fei-Fei, L.}
\newblock Large-scale video classification with convolutional neural networks.
\newblock In {\em Proceedings of the IEEE conference on Computer Vision and
  Pattern Recognition\/} (2014), pp.~1725--1732.

\bibitem{Kearnes2016}
{\sc Kearnes, S., McCloskey, K., Berndl, M., Pande, V., and Riley, P.}
\newblock {Molecular graph convolutions: moving beyond fingerprints}.
\newblock {\em J. Comput.-Aided Mol. Des. 30}, 8 (2016), 595--608.

\bibitem{kingma2014adam}
{\sc Kingma, D., and Ba, J.}
\newblock Adam: A method for stochastic optimization.

\bibitem{kipf2016semi}
{\sc Kipf, T.~N., and Welling, M.}
\newblock Semi-supervised classification with graph convolutional networks.
\newblock {\em arXiv preprint arXiv:1609.02907\/} (2016).

\bibitem{kleinberg2006algorithm}
{\sc Kleinberg, J., and Tardos, E.}
\newblock {\em Algorithm design}.
\newblock Pearson Education India, 2006.

\bibitem{krizhevsky2012imagenet}
{\sc Krizhevsky, A., Sutskever, I., and Hinton, G.~E.}
\newblock Imagenet classification with deep convolutional neural networks.
\newblock In {\em Advances in neural information processing systems\/} (2012),
  pp.~1097--1105.

\bibitem{lecun2015deep}
{\sc LeCun, Y., Bengio, Y., and Hinton, G.}
\newblock {Deep learning}.
\newblock {\em Nature 521}, 7553 (may 2015), 436--444.

\bibitem{lengauer2012combinatorial}
{\sc Lengauer, T.}
\newblock {\em Combinatorial algorithms for integrated circuit layout}.
\newblock Springer Science \& Business Media, 2012.

\bibitem{li2015gated}
{\sc Li, Y., Tarlow, D., Brockschmidt, M., and Zemel, R.}
\newblock Gated graph sequence neural networks.
\newblock {\em arXiv preprint arXiv:1511.05493\/} (2015).

\bibitem{magnanti1984network}
{\sc Magnanti, T.~L., and Wong, R.~T.}
\newblock Network design and transportation planning: Models and algorithms.
\newblock {\em Transportation science 18}, 1 (1984), 1--55.

\bibitem{Matlock2018}
{\sc Matlock, M.~K., Dang, N.~L., and Swamidass, S.~J.}
\newblock Learning a local-variable model of aromatic and conjugated systems.
\newblock {\em {ACS} Central Science 4}, 1 (jan 2018), 52--62.

\bibitem{Matlock2015}
{\sc Matlock, M.~K., Hughes, T.~B., and Swamidass, S.~J.}
\newblock {XenoSite server: A web-available site of metabolism prediction
  tool}.
\newblock {\em Bioinformatics 31}, 7 (2015), 1136--1137.

\bibitem{mikolov2013efficient}
{\sc Mikolov, T., Chen, K., Corrado, G., and Dean, J.}
\newblock Efficient estimation of word representations in vector space.
\newblock {\em arXiv preprint arXiv:1301.3781\/} (2013).

\bibitem{milletari2016v}
{\sc Milletari, F., Navab, N., and Ahmadi, S.-A.}
\newblock V-net: Fully convolutional neural networks for volumetric medical
  image segmentation.
\newblock In {\em 3D Vision (3DV), 2016 Fourth International Conference on\/}
  (2016), IEEE, pp.~565--571.

\bibitem{niepert2016learning}
{\sc Niepert, M., Ahmed, M., and Kutzkov, K.}
\newblock Learning convolutional neural networks for graphs.
\newblock In {\em International conference on machine learning\/} (2016),
  pp.~2014--2023.

\bibitem{o2011open}
{\sc O'Boyle, N.~M., Banck, M., James, C.~A., Morley, C., Vandermeersch, T.,
  and Hutchison, G.~R.}
\newblock Open babel: An open chemical toolbox.
\newblock {\em Journal of cheminformatics 3}, 1 (2011), 33.

\bibitem{page1998pagerank}
{\sc Page, L., Brin, S., Motwani, R., and Winograd, T.}
\newblock The pagerank citation ranking: Bringing order to the web.

\bibitem{pereira2016brain}
{\sc Pereira, S., Pinto, A., Alves, V., and Silva, C.~A.}
\newblock Brain tumor segmentation using convolutional neural networks in mri
  images.
\newblock {\em IEEE transactions on medical imaging 35}, 5 (2016), 1240--1251.

\bibitem{Perozzi2014}
{\sc Perozzi, B., Al-Rfou, R., and Skiena, S.}
\newblock {DeepWalk}.
\newblock In {\em Proceedings of the 20th {ACM} {SIGKDD} international
  conference on Knowledge discovery and data mining - {KDD}
  {\textquotesingle}14\/} (2014), {ACM} Press.

\bibitem{raymond2002maximum}
{\sc Raymond, J.~W., and Willett, P.}
\newblock Maximum common subgraph isomorphism algorithms for the matching of
  chemical structures.
\newblock {\em Journal of computer-aided molecular design 16}, 7 (2002),
  521--533.

\bibitem{ronneberger2015u}
{\sc Ronneberger, O., Fischer, P., and Brox, T.}
\newblock U-net: Convolutional networks for biomedical image segmentation.
\newblock In {\em International Conference on Medical image computing and
  computer-assisted intervention\/} (2015), Springer, pp.~234--241.

\bibitem{PySpice}
{\sc Salvaire, F.}
\newblock Pyspice.
\newblock \url{https://pyspice.fabrice-salvaire.fr}, 2018.

\bibitem{scarselli2009graph}
{\sc Scarselli, F., Gori, M., Tsoi, A.~C., Hagenbuchner, M., and Monfardini,
  G.}
\newblock The graph neural network model.
\newblock {\em IEEE Transactions on Neural Networks 20}, 1 (2009), 61--80.

\bibitem{Schmidhuber2015}
{\sc Schmidhuber, J.}
\newblock {Deep learning in neural networks: An overview}.
\newblock {\em Neural Networks 61\/} (2015), 85--117.

\bibitem{scott2017social}
{\sc Scott, J.}
\newblock {\em Social network analysis}.
\newblock Sage, 2017.

\bibitem{shibata2008detecting}
{\sc Shibata, N., Kajikawa, Y., Takeda, Y., and Matsushima, K.}
\newblock Detecting emerging research fronts based on topological measures in
  citation networks of scientific publications.
\newblock {\em Technovation 28}, 11 (2008), 758--775.

\bibitem{shore1981properties}
{\sc Shore, J., and Johnson, R.}
\newblock Properties of cross-entropy minimization.
\newblock {\em IEEE Trans. Inf. Theory 27}, 4 (1981), 472--482.

\bibitem{Szklarczyk2014}
{\sc Szklarczyk, D., Franceschini, A., Wyder, S., Forslund, K., Heller, D.,
  Huerta-Cepas, J., Simonovic, M., Roth, A., Santos, A., Tsafou, K.~P., Kuhn,
  M., Bork, P., Jensen, L.~J., and von Mering, C.}
\newblock {STRING} v10: protein{\textendash}protein interaction networks,
  integrated over the tree of life.
\newblock {\em Nucleic Acids Research 43}, D1 (oct 2014), D447--D452.

\bibitem{tai2015improved}
{\sc Tai, K.~S., Socher, R., and Manning, C.~D.}
\newblock Improved semantic representations from tree-structured long
  short-term memory networks.
\newblock {\em arXiv preprint arXiv:1503.00075\/} (2015).

\bibitem{tang2015line}
{\sc Tang, J., Qu, M., Wang, M., Zhang, M., Yan, J., and Mei, Q.}
\newblock Line: Large-scale information network embedding.
\newblock In {\em Proceedings of the 24th international conference on world
  wide web\/} (2015), International World Wide Web Conferences Steering
  Committee, pp.~1067--1077.

\bibitem{vishwanathan2010graph}
{\sc Vishwanathan, S. V.~N., Schraudolph, N.~N., Kondor, R., and Borgwardt,
  K.~M.}
\newblock Graph kernels.
\newblock {\em Journal of Machine Learning Research 11}, Apr (2010),
  1201--1242.

\bibitem{zaretzki2013xenosite}
{\sc Zaretzki, J., Matlock, M., and Swamidass, S.~J.}
\newblock Xenosite: accurately predicting cyp-mediated sites of metabolism with
  neural networks.
\newblock {\em Journal of chemical information and modeling 53}, 12 (2013),
  3373--3383.

\bibitem{zerbino2008velvet}
{\sc Zerbino, D.~R., and Birney, E.}
\newblock Velvet: algorithms for de novo short read assembly using de bruijn
  graphs.
\newblock {\em Genome research 18}, 5 (2008), 821--829.

\end{thebibliography}


\begin{thebibliography}{1}

\bibitem{bengio2009curriculum}
{\sc Bengio, Y., Louradour, J., Collobert, R., and Weston, J.}
\newblock Curriculum learning.
\newblock In {\em Proceedings of the 26th annual international conference on
  machine learning\/} (2009), ACM, pp.~41--48.

\bibitem{Matlock2018}
{\sc Matlock, M.~K., Dang, N.~L., and Swamidass, S.~J.}
\newblock Learning a local-variable model of aromatic and conjugated systems.
\newblock {\em {ACS} Central Science 4}, 1 (jan 2018), 52--62.

\bibitem{quarles1989analysis}
{\sc Quarles, T.~L.}
\newblock Analysis of performance and convergence issues for circuit
  simulation.
\newblock Tech. rep., California University Berkeley Electronics Research Lab,
  1989.

\bibitem{PySpice}
{\sc Salvaire, F.}
\newblock Pyspice.
\newblock \url{https://pyspice.fabrice-salvaire.fr}, 2018.

\end{thebibliography}

\end{document}

\maketitle

\section{Supplementary methods}

\subsection{miniGRU}

The miniGRU, adapted from Matlock \textit{et al}~\cite{Matlock2018}, is a variant of the gated recurrent unit in which the read gate is omitted.  Formally, given a input $x_{p-1}$ and memory state $s_{p-1}$, the recurrence equations for pass $p$ are:

\begin{equation}
    \begin{split}
        u_p & = \sigma( W_1 x_{p-1} + W_2 s_{p-1} + b_1 ), \\
        o_p & = f(W_3 x_{p-1} + W_4 s_{p-1} + b_2), \\
        s_p & = (1-u_p) \odot s_{p-1} + u_p \odot o_p,
    \end{split}
\end{equation}

where $\odot$ is element-wise vector multiplication, $\sigma$ is the sigmoid activation function, $f$ is the exponential linear activation function, the $W_*$ are weight matrices and the $b_*$ are bias vectors. The output state $s_p$ is both the output to the next layer and the input to state for the next recurrence.

\subsection{Curriculum learning}

We adopt a curriculum learning approach to the path finding, maze solving and circuit analysis problems. To construct the curriculum, training examples are sorted into bins based on their size (number of nodes), with bin $1$ being the smallest examples ($4 \times 4$ nodes). We then define a probability distribution $p_i$ over training example bins $i$. At each iteration of training, a minibatch is drawn from a bin with probability $p_i$. At the beginning of training, $p_1=1$ and $p_i=0$ for all $i>1$. Let $\phi$ be the index of the first bin with $p_\phi=0$. After a fixed number of iterations, the bin probabilities are updated as:

\begin{equation}
    p_i = p_i\eta + \frac{1-\eta}{\phi},
\end{equation}

for all $i \leq \phi$, where $\eta$ is a decay factor ($0.25$ in our experiments). All the $p_i$ converge to $1/N$, where $N$ is the number of bins. The entropy of this distribution increases monotonically, in accordance with the definition of curriculum learning~\cite{bengio2009curriculum}.

\subsection{Computing voltages in a circuit}

To construct random circuits for this problem, we started with a grid graph of $N \times N$ nodes. Each edge was subject to random deletion with probability $d$, which varied by circuit size (Table~\ref{tab:circuit_training_data}). For each remaining edge, a component was placed on that edge with the following probabilities: battery $0.05$, resistor $0.7$, wire $0.25$. Batteries were placed with a random orientation of positive and negative terminals. If a battery was not present at the end of this procedure, a random edge was chosen and replaced with a battery. Properties for each component were also chosen randomly. For resistors, resistance was chosen from a uniform distribution between \SI{100}{\ohm} and \SI{1000}{\ohm}. For batteries, voltage was chosen from a uniform distribution between \SI{5}{\volt} and \SI{20}{\volt}. All batteries were given an internal resistance of \SI{100}{\ohm} to ensure voltages and currents were well defined for all circuits. Operating point voltages were computed with PySpice~\cite{PySpice}, an interface to the commonly used Spice circuit simulation library~\cite{quarles1989analysis}. Training examples were generated with sizes between two and $10$. For each size, $500$ training mini-batches were generated. Batch sizes decreased with sample size to accommodate the increased computation time required by the circuit analysis software (Table~\ref{tab:circuit_training_data}). A single batch of $100$ examples was generated for test data at each grid size from $2$ to $15$. The same delete frequencies were used as defined in the training data. The delete frequency for test examples with sizes $11$ to $15$ was $0.5$.

We converted PySpice circuits to graph examples for input to wave and graph convolution networks in four steps. First, (1) we labeled nodes with their voltages compared to ground (training targets) as determined by circuit analysis software PySpice. Second, (2) we labeled edges with two properties: resistance and voltage. If an edge was a resistor it was labeled with $0$ voltage. If an edge was a wire, it was labeled with $0$ voltage and $0$ resistance. Only battery edges had nonzero voltage. Third, (3) we labeled nodes connected to the positive terminal of a battery to indicate battery orientation. A temperature encoding was used to indicate up to four batteries connected to the same node. Finally, (4) we converted the edge properties (battery voltage and resistance) to a log space by the transformation $\log(1+x$).

\begin{table}
    \caption{Training data parameters for circuit analysis problem}
  \centering
    \label{tab:circuit_training_data}
\begin{tabular}{ccc}
\toprule
    Grid size & Edge deletion probability (d) & Batch size \\
\midrule
    2 & 0.1 & 100 \\
    3 & 0.1 & 90 \\
    4 & 0.2 & 80 \\
    5 & 0.2 & 70 \\
    6 & 0.3 & 60 \\
    7 & 0.4 & 50 \\
    8 & 0.5 & 40 \\
    9 & 0.5 & 30 \\
    10 & 0.5 & 20 \\
\bottomrule
\end{tabular}
\end{table}

\newpage

\bibliographystyle{acm}
\bibliography{graph_deeplearning}